# Topology Control of wireless sensor network using Quantum Inspired Genetic algorithm

Sajid Ullah
Institute of space technology, Pakistan
sajidullah.ist@gmail.com

Mussarat Wahid
University of Nottingham
mussaratwahid@yahoo.com

*Abstract*—In this work, an evolving Linked Quantum Register has been introduced, that are group vector of binary pair of genes, uses local topological space to represent those nodes. The optimal points of node for topology control have high connectivity and have low energy consumption, and have interference at low. The register works in higher dimension. In this modeling order-2 Quantum Inspired Genetic Algorithm has been used and higher order can be used to achieve greater versatility in topology control of nodes. Numerical result has been obtained, analysis is done as how the result has previously been obtained with Quantum Genetic Algorithm and results are compared too. For future work, metrics for LQR are hinted that would exploit the algorithm to work in more computational intensive problem.

*Key Words— relative quantum order, topology control, Linked quantum register*

## I. INTRODUCTION

The two algorithms, evolutionary and quantum computation has made a remarkable success by solving some problem with relatively less computational time and memory. From it a new meta heuristics field has been made that combined the power of two. Genetic programming has been used in number of application. [13], [14], and [15] The work, Quantum inspired genetic algorithm (QIGA) was put forward by Narayan in 1996 [1]. Although evolutionary techniques combined with quantum computing has been for years but the first work on practical problem to work out was used for combinatory optimization [2]. Quantum inspired genetic algorithm used probability based optimization that include algorithms from evolutionary and quantum computing with the likes of parallel computing, superposition and genetic variation [1].QIGA does not require quantum computer for its implementation, but there are metrics when reached certain level, then can quantum computer be required. Those computing techniques combine with probabilities achieve such great result that converge for global optimum. QIGA has been used for combinatory optimization problem, and solved several computational intensive problems. QIGA has been used in Power system optimization and localization of mobile robots [3], Image processing [4], flow shop Scheduling [5], Optimization of Hot
 Extrusion Process [6]. Several algorithms are used for topology control in wireless network (WSN). LEACH [12], SPAN [7], ASCENT [8], and STEM [9]. The first use of evolutionary and Quantum computing used for topology control was in the work [10], that show better result than simple genetic algorithm. The topology control was achieved in fewer generation in Quantum genetic algorithm (QGA). QGA proves superior to simple genetic algorithm. Higher order QIGA was presented in work [11]. Order-2 QIGA does need a functional quantum computer for its implementation as it uses random features from quantum mechanical system such as qubits, superposition, and parallel computing. Quantum factor defines when algorithm have to be implemented on quantum computer, and is elaborated in [11]

The optimal point of topology control is standardized on the connectivity of nodes, number of nodes exceeding the threshold, energy of nodes, and interference between the nodes. The interference is one of key issue in wireless sensor network (WSN). An optimal point is selected that minimize the interference in the register and the intra node between the registers. Interference has not been accounted in topology control here. Higher order QIGA have terminology that does not exist in QGA. Quantum and relative quantum order, and quantum factor are defined here and for detailed work, the reader should understand the full algorithm in [11].

This paper is structured as: section 1 will describe the various design metrics, that influence the topology of nodes. Section 2 discussed the order -2 QIGA, various definitions of the algorithm are introduced. In section 3 Linked quantum register has been introduced. In section 4 algorithm has been implemented for topology control. In Last section,

Numerical result has been obtained for topological control of nodes.

## II. TOPOLOGICAL INITIALIZATION, CONSTRUCTION AND MAINTENANCE

The sensor network is initialized with **n** number of sensor node at t=0, all the nodes have equal resources.

So at the start of the process the nodes have identical probability amplitudes that implies from the above statement, initially the nodes are randomly selected in topological space in that they are sensing, surveillance or other data acquisition. The second necessary step in topological control is its construction for that we present a novel higher dimensional register LQR, that varies itself across search space and Migration status is updated on each iteration. The Topology maintenance is what the algorithm does in a more optimal way than QIGA, On each successive iteration the topological point in space between nodes are varied as to keep the nodes net energy at a minimal level, maintain the connectivity of node strong enough so that don't interfere in one another wireless information signal. In wireless communication control decision are most energy consuming, that are kept to minimum number.

Suppose there are **n** number of nodes in a network, $1 \leq i \leq n$, that in coordinate space are located at $(x_i, y_i)$ at the radius of $r_i$. The power consumes by each for any node is $P_i$.

$R_{max}$ –the maximum a node can transmit

$A_i$ - show the connectivity of the network, considering it, the network has to ensure it doesn't interfere in adjacent node' transmission region.

$R_f$ – the nodes that exceed the threshold point $R_t$ are $b_k$ in number.

The topological factors mathematically are

$\forall C_{ij}$ in $C^{(n)}$, n defines the dimensionality of the node.

The element of adjacency matrix is one if they are connected, that should be in a cluster for node connected in a register. For topology control WSN attains these points.

- The total energy of the node should keep at a minimum
- The number of nodes that exceeds the threshold are to be sparse.
- The interference between the elements of adjacent register in topological space should be lowered.

## III. ORDER-2 QUANTUM INSPIRED GENETIC ALGORITHM

The QIGA uses individual independent qubits to represent binary genes, so they are effectively in order 1. Higher order QIGA was presented in [11]. We here present introductory definition from their work. For full understanding of their work reader should study [11]. QIGA-2 symbolizes classical and quantum individual population by *P* and *Q*. Order *r* is defined as the largest register used in the algorithms. $1 \leq r \leq N$.

- Quantum order **r** is the size of biggest quantum register used in the algorithm.
- Relative quantum order **w** shows the ratio of quantum order to problem size, N. For QIGA-2, the 100 individual qubits would have a Relative Quantum range of $\frac{2}{100} = 0.02$, This is representation based on 2-qubit registers
- Quantum factor ($\lambda \in [0\ 1]$) that define the ratio of the ratio of the dimension of space of a current Algorithm to the dimension of space of the full quantum register of all individuals.

$$\lambda = \frac{2^r}{w2^N}$$

In our topological control problem, there can be a scenario in a high dimensional space in that there will be a need to deploy different registers size for nodes in local proximity of topological space. In QIGA-2, register have a pair of nodes.

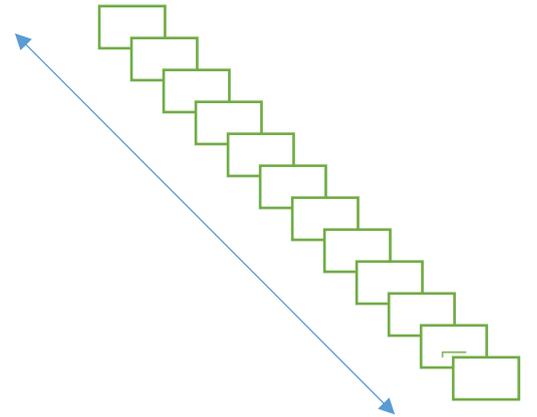

(a) Chromosome in order-1 QIGA

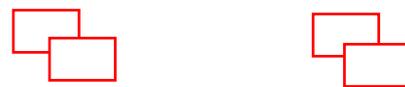

(b) Chromosome in Order-2 QIGA

Fig. 1. (a): In order one each individual chromosome is spanned in two dimensional space. (b): In Order-2 the register has two chromosomes, and dimension of space for the order-2 QIGA is 4

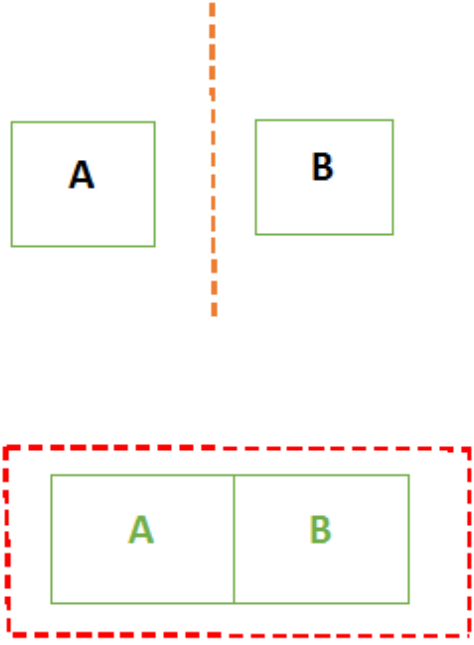

Fig. 2. The 1st depiction shows two nodes in an unlocked manner while the second show nodes in a Quantum locked register.

## IV. LINKED QUANTUM REGISTER (LQR)

In this work we propose a model that takes the adjacent 2-qubit quantum register into account to form a binary paired register. The pair of binary genes in a register represent a node, while in register have a pair of nodes. The information between nodes in a register is transferred that achieve the topology optimal point in space. As a result, the LQR has four points in probability in distribution. $|\alpha_0|^2$, $|\alpha_1|^2$, $|\alpha_2|^2$, and $|\alpha_3|^2$.

These are a few key points about LQR.

- There can be an arbitrary number of Quantum binary genes in the register. It can vary across different register depending upon the computational complexity a classical machine can cope with. The network can have different number of chromosomes in a register, and still can achieve the work of having the same number of chromosomes (nodes) with turning off those nodes for the iterative procedure, but survival of the fittest results from evolutionary computing will keep the nodes ahead in the run.
- The protocol used are application specific in a register. The control decision information on the elements of a register is specific to the individuals only
- LQR has memory because in each update step, that is the distance of nodes in each Register, is to be stored.

### a. Algorithm for Topology Control

The QIGA-2 starts by searching the adjacent nodes for that the register identifies (RI) is identical. After selecting the nodes of same RI. The algorithm checks for the node connectivity in adjacency matrix that is updated by algorithm presented in [11]. The connectivity of nodes is range of space between the nodes or discrete set of point in space. A prior optimal point is set for WSN that is application dependent. A loop is run over the node and the distance is incremented or decremented between the node is achieved that assign node to the same RI. The algorithms then compute the distance between the nodes and set it to be connected. The position of each node is varied by choosing the number of steps in each iterative process, that results in step size of the increment or decrement of topology control. The algorithm converges when the given node in WSN is in the connectivity range set prior in the algorithm. The update of 2x2 dimension of LQR depends by choosing one or both of the node. The algorithm has **select** () method that chooses one of the nodes in QLR.

**Algorithm 1:** Topology Control in QIGA2

1: **for** i=0,1,2,...,N
2: **for** j=0,1,2,...,N-1/2
3: $dist_{conn}(i,i+1)=1$
4: **select( i,i+1)** for j=j+1
5: $dist(i,i+1)$
6: **if** $dist(i,i+1)=dist_{conn}(i,i+1)$
7: **then** Mat[i,i+1]=1
8: **else if** $dist(i,i+1) \leq dist_{conn}(i.i+1)=1$
9: **while** $dist(I,i+1) \neq dist_{conn}(i,i+1)$
10: **do** $dist(i,i+1) \leftarrow dist(i,i+1)+\Delta i$
11: **else if** $dist(i,i+1) \geq dist_{conn}(i,i+1)$
12: **while** $dist(I,i+1) \neq dist_{conn}(i,i+1)$
13: **do** $dist(i,i+1) \leftarrow dist(i,i+1)-\Delta i$
14: **end for**
15: **end for**

## V. ALGORITHM IMPLEMENTATION FOR TOPOLOGY CONTROL

Several algorithms have been implemented for topology control [7], [8], [9], [10], and [12].
The algorithm starts with an arbitrary number of wireless sensor network (1 ≤ i ≤ n). The initial population will start with equal resources in term of energy, processing Capability and Transceiver power.
In order-2 QIGA the population is represented by a vector of chromosome (register). All amplitudes are initialized with a magnitude of $\frac{1}{\sqrt{2}}$

$$Q(t) = \{q_1^t, q_2^t ... q_n^t\}$$

**Solution of individual Q-bits in register:**

In this step, binary string is produced from qubits string, this is called observation of states. Q (t) with n bits represent a superposition of $2^n$ binary states of gene. The states of individual in the register are observed
For any Register, $R_{id}$
$$P(t) = \{p_1^t, p_2^t, …, p_n^t\}$$
Where $P^t_j$ represent observational result of $j^{th}$ individual, $R_{id}$ will have N/r number of P (t) in each iterative step
In evaluation step, LQR has the individual nodes, and they are evaluated on the basis of how strongly they are connected. The individual use minimum energy and minimum number of nodes that exceeds the power transmission threshold radius, $R_f$. The Binary solution is stored.
$$B(t) = \{b_1^t, b_2^t, …, b_n^t\}$$
Then the observation of state of Q (t-1) produce a binary solution in P (t). The variation operator is model specific, and the user can pick any of the operators that suits the algorithm the best. In order-2, genetic operators are taken 4x4 quantum gates. The observation function from step 2 of algorithm returns strings of binary genes 00, 01, 10 and 11 with a probability of $|α_0|, |α_1|, |α_2|$, and $|α_3|$ respectively.
Topology was obtained by
The modeling of LQR makes the difference in how topology is obtained unlike QGA. The pair of nodes in LQR works in coordination checking the node in its LQR and nodes in adjacent LQR for the topological condition the algorithm applies.

## VI. NUMERICAL RESULTS

The algorithm has been implemented for 16 nodes. The WSN took 79 generations to accomplish the condition specified in algorithm to converge. Roulette wheel was used for the evaluation function of fitness. The QIGA-2 has been compared with Simple Quantum Genetic (SGA) algorithm. Solution found in QIGA-2 earlier, that was obtained because there were two pair of LQR that were in transmission range of one another. So as a result, it takes fewer generation for finding the optimal topology control.

Table 1. Result of algorithm: QGA and QIGA-2

| Algorithms | Earliest generation finding the solution | Solution found in the generation | Total power the radius threshold | Rf |
|---|---|---|---|---|
| QGA | 90 | 47 | 156 | 7 |
| QIGA-2 | 79 | 42 | 140 | 6 |

LQR modeling in QIGA-2 results in less generation for topology control than QGA. The solution obtained earlier because the two pair of LQR that have transmission region exceeded than standardized earlier. The power consumption due to less generation of WSN was beneficial in QIGA-2, and results in less net power than QGA.

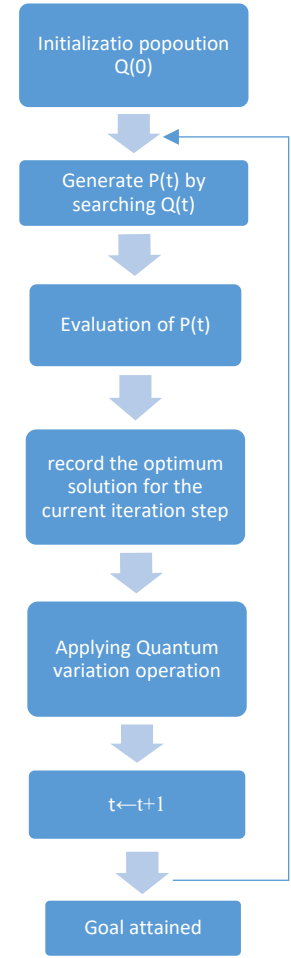

Fig 3. Updating Quantum gene state space

## VII. CONCLUSION

In this work, a novel register is introduced based on QIGA-2 that uses evolutionary computing in addition to quantum computing to find the optimal solution for topology control. The attributes inherited from quantum computing in LQR are the way of representation of solution of nodes. The order 2 register use the nodes in register to cooperate in the information necessary to increase the connectivity to an optimal level. At the start of algorithm, nodes are paired to make LQR, that transfer the information in each iteration of the algorithm for the topology control. Future perspective of the QIGA lies in using more than two number of node in LQR that seem to obtain better result. Modeling of LQR for more than two nodes and its implementation on Quantum computer would be challenging.

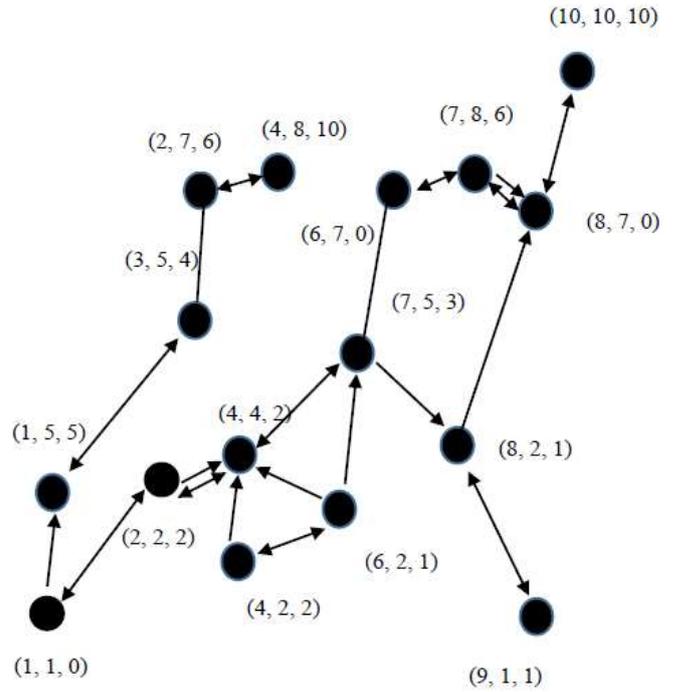

Fig 4. Topology attained by QIGA-2

ACKNOWLEDGMENT

We would like to thank our colleagues from institute of space technology and university of Nottingham for their valuable discussion about the problem and insights that greatly assists in our research work.